\documentclass{article}

\usepackage{arxiv}

\usepackage[utf8]{inputenc} 
\usepackage[T1]{fontenc}    
\usepackage{hyperref}       
\usepackage{url}            
\usepackage{booktabs}       
\usepackage{amsfonts}       
\usepackage{nicefrac}       
\usepackage{microtype}      
\usepackage{lipsum}         
\usepackage{graphicx}
\usepackage{natbib}
\usepackage{doi}

\usepackage{amsmath,amssymb,amsfonts}
\usepackage{algorithmic}
\usepackage{graphicx}
\usepackage{textcomp}

\usepackage[table]{xcolor}
\usepackage{adjustbox}
\usepackage{rotating}
\usepackage{multirow}
\usepackage{makecell}
\usepackage{hyperref}
\usepackage{longtable} 
\usepackage{colortbl}  
\usepackage{xtab}
\usepackage{listings}
\usepackage{comment}
\usepackage{array,booktabs,makecell}
\usepackage{pict2e}
\usepackage{url}

\usepackage{breakurl}

\usepackage{subcaption}
\usepackage{geometry}
\usepackage{cleveref} 

\newif\ifshowchanges
\showchangesfalse  

\usepackage{xcolor}
\usepackage{soul} 
\usepackage[normalem]{ulem}
\ifshowchanges
  \newcommand{\added}[1]{\textcolor{blue}{#1}}
\newcommand{\deleted}[1]{\textcolor{red}{\sout{#1}}}
\newcommand{\replaced}[2]{\textcolor{purple}{\sout{#1}}\textcolor{blue}{#2}}
\else
  \newcommand{\added}[1]{#1}  
  \newcommand{\deleted}[1]{}  
  \newcommand{\replaced}[2]{#2}  
\fi

\title{The Effectiveness of Large Language Models in Transforming Unstructured Text to Standardized Formats}

\date{}

\newif\ifuniqueAffiliation
\uniqueAffiliationtrue

\ifuniqueAffiliation 
\author{ 
\hspace{1mm}William Brach\\
	Slovak Technical University\\
	\texttt{william.brach@stuba.sk} \\
	\And
	\hspace{1mm}Kristi\'{a}n~Ko\v{s}\v{t}\'{a}l \\
	Slovak Technical University\\
	\texttt{kristian.kostal@stuba.sk} \\
    	\And
	\hspace{1mm}Michal Ries \\
	Slovak Technical University\\
	\texttt{michal.ries@stuba.sk} \\
}

\renewcommand{\shorttitle}{\textit{arXiv} Template}

\begin{document}
\maketitle

\begin{abstract}
The exponential growth of unstructured text data presents a fundamental challenge in modern data management and information retrieval. While Large Language Models (LLMs) have shown remarkable capabilities in natural language processing, their potential to transform unstructured text into standardized, structured formats remains largely unexplored - a capability that could revolutionize data processing workflows across industries. This study breaks new ground by systematically evaluating LLMs' ability to convert unstructured recipe text into the structured Cooklang format. Through comprehensive testing of four models (GPT-4o, GPT-4o-mini, \replaced{Llama3.1:70b}{Llama3.3:70b}, and Llama3.1:8b), an innovative evaluation approach is introduced that combines traditional metrics (WER, ROUGE-L, TER) with specialized metrics for semantic element identification. Our experiments reveal that GPT-4o with few-shot prompting achieves breakthrough performance (ROUGE-L: 0.8209, WER: 0.3509), demonstrating for the first time that LLMs can reliably transform domain-specific unstructured text into structured formats without extensive training. Although model performance generally scales with size, we uncover surprising potential in smaller models like Llama3.1:8b for optimization through targeted fine-tuning. These findings open new possibilities for automated structured data generation across various domains, from medical records to technical documentation, potentially transforming the way organizations process and utilize unstructured information.

For further information, source code, and associated resources, please refer to the source code repository\footnote{\url{https://github.com/williambrach/llm-text2cooklang}}.

\end{abstract}

\section{Introduction}

The exponential growth of unstructured text data presents a challenge in data management, analysis, and information retrieval. While Large Language Models (LLMs) \cite{brown2020languagemodelsfewshotlearners} have demonstrated remarkable capabilities in natural language processing, their potential to transform unstructured text into standardized, structured domain-specific formats remains unexplored. This capability could revolutionize data processing workflows across various domains, enabling more efficient information retrieval, automated analysis, and data management.

This study addresses these gaps by examining the potential of LLMs in structured text generation \cite{raffel2023exploringlimitstransferlearning}, specifically focusing on converting plain text recipes into Cooklang specifications \cite{cooklang_2024}. In this study, we aim to address two research questions: How do different LLM architectures affect the quality and accuracy of structured recipe generation? What impact do various prompting techniques \cite{liu2021pretrainpromptpredictsystematic} have on the generation of well-formed Cooklang specifications? The applications of this research extend beyond the culinary domain, with potential implications for healthcare documentation using HL7 format \cite{dolin2001hl7}, music notation in MusicXML \cite{good2012musicxml}, and technical documentation organization.

In order to evaluate the performance of LLM systems in the generation of structured recipes, a variety of methods are employed for both qualitative and quantitative assessment. Our methodology included traditional natural language processing metrics such as Word Error Rate (WER) \cite{wer_ali2018word,wer_klakow2002testing}, and ROUGE-L \cite{google2019rouge,google_see2017pointsummarizationpointergeneratornetworks}, which measure the accuracy and fluency of the generated text. Additionally, we introduce domain-specific metrics to evaluate the accuracy of a generated recipe. These include an \textit{Ingredient Identification Score} and \textit{Unit and Amount Identification Scores}. The evaluation of these metrics allows for the assessment of not only the linguistic quality of the generated recipes but also their adherence to the specific structural requirements of the Cooklang format and their culinary accuracy. Furthermore, we analyze the LLMs performance across different input configurations and prompting techniques (zero-shot \cite{promptingguide_zeroshot}, few-shot \cite{promptingguide_fewshot}, and MIPROv2 \cite{opsahlong2024optimizinginstructionsdemonstrationsmultistage}). This evaluation approach enables us to understand the strengths and limitations of various LLMs in the context of structured domain-specific format generation.

\added{In this study, we aim to address two research questions: How do different LLM sizes affect the quality and accuracy of structured recipe generation? What impact do various prompting techniques have on the generation of well-formed Cooklang specifications? The applications of this research extend beyond the culinary domain, with potential implications for healthcare documentation using HL7 format, music notation in MusicXML, and technical documentation organization. The key contributions of this research include:}

\added{\begin{itemize}
    \item A systematic evaluation of LLMs' ability to transform unstructured recipe text into structured Cooklang format across different models and sizes. 
    \item Identification of scaling patterns across model sizes, with the surprising finding that smaller models like Llama3.1:8b show potential for optimization through targeted fine-tuning. 
    \item Demonstration that LLMs can reliably transform domain-specific unstructured text into structured formats without requiring extensive training. 
    \item Empirical evidence that GPT-4o with few-shot prompting achieves breakthrough performance (ROUGE-L: 0.8209, WER: 0.3509) in structured format conversion. 
    \item A framework for implementing domain-specific evaluation methodology that combines traditional metrics (WER, ROUGE-L, TER) with specialized metrics for semantic element identification in recipes. 
    \item Insights that open new possibilities for automated structured data generation across various domains, including healthcare, technical documentation, and legal contracts 
\end{itemize}}

The remainder of this paper is organized as follows: Section \ref{background} reviews related work in structured text generation and prompting techniques. Section \ref{methodology} describes our methodology and evaluation framework. Section \ref{results} presents our experimental results and examination. Section \ref{discussion} discusses the implications and limitations of our findings, and Section \ref{sec:conclusion} concludes with future research directions.

\section{Background}
\label{background}

The standardization of recipe processing and cooking knowledge has evolved considerably since Mundie's pioneering work \cite{mundie1985computerized} in 1985, which introduced RxOL (Recipe Operation Language) as one of the earliest structured approaches to computational recipe representation. In web search, structured data specification \cite{google_recipe_structured} for recipes represents a significant modern standardization effort, providing a systematic way to encode recipe information for search engines and enabling enhanced discovery and presentation of recipe content across the web. This work laid the foundation \cite{alexey2015cooking} for subsequent advancements in the field, as evidenced by research that drew innovative parallels between software design patterns and cooking procedures, suggesting that cooking patterns could serve as an abstraction layer for standardizing recipe knowledge. The development of domain-specific markup languages (e.g., Cooklang for recipes \cite{cooklang_2024}, MusicXML \cite{good2012musicxml}for digital sheet music, and HL7 \cite{dolin2001hl7} for healthcare data) has emerged to standardize information representation in specific domains. These markdown languages offer a structured approach to encoding domain knowledge, enhancing readability and processing efficiency. There has been an increase in the number of initiatives \cite{bommasani2021opportunities,pitsilou2024using,rostami2024integrated} focused on leveraging LLMs in the culinary domain or other domains that could leverage this structured approach in free texts. 

One of the use cases for LLMs is the extraction of food entities from cooking recipes, as demonstrated in the study by Pitsilou et al. \cite{pitsilou2024using}. The results of this study demonstrate that, even in the absence of labeled training data, LLMs can achieve promising results in domain-specific tasks such as named entity recognition (NER) for food items. Another illustration of the application of LLMs for structured text parsing outside the culinary domain is StructGPT, a general framework to enhance LLMs' zero-shot reasoning capabilities over structured data such as knowledge graphs, tables, and databases \cite{jiang2023structgpt}. An additional example of the use of LLMs in a culinary context is LLaVA-Chef, a multi-modal generative model specifically designed for food recipes \cite{mohbat2024llava}. By adapting the LLaVA (Large Language and Vision Assistant) \cite{liu2024improvedbaselinesvisualinstruction} model to the food domain through a multi-stage fine-tuning approach, the authors achieved state-of-the-art performance in recipe generation tasks. Notably, these results demonstrate competitive performance in comparison to the work of Patil et al. \cite{Pansare2024ARO}, who employed convolutional neural networks (CNNs) for image recognition in conjunction with deep learning models for natural language processing. Another illustration of the application of LLMs to culinary tasks is the development of RecipeNLG \cite{bien2020recipenlg}, which employed language models to enhance semi-structured recipe text generation and generate a dataset for subsequent LLM training and refinement. As this domain progresses, it demonstrates the potential for integrating general-purpose language models with specialized domain expertise. Building on this work, Zhou et al.'s research on FoodSky \cite{zhou2024foodskyfoodorientedlargelanguage} demonstrated that LLMs can achieve better performance in domain-specific tasks by using specialized mechanisms like the Topic-based Selective State Space Model (TS3M) and Hierarchical Topic Retrieval Augmented Generation (HTRAG). The study demonstrated the effectiveness of FoodSky by achieving accuracy rates of 67.2\% and 66.4\% in chef and dietetic examinations, respectively. Authors suggest that LLMs, when adequately trained and enhanced with domain-specific knowledge, can efficiently handle structured text conversion tasks. This lays a solid foundation for applications in recipe processing and other specialized fields. The research paper Cook2LTL \cite{10611086} demonstrates how to utilize a language model to translate recipes into linear temporal logic (LTL). Cook2LTL shows how structured representations can bridge the gap between human-readable content and machine-executable instructions. Another application of LLM \cite{dubovskoy_ai_evolution} employs graph-based representations in order to track ingredient transformations and state changes throughout the cooking process. This evolution, from basic formalization to pattern-based approaches and finally to LLM-powered graph representations, illustrates the ongoing efforts to bridge the gap between natural language cooking instructions and structured, computationally accessible formats. 

\added{Study "Fine-tuning Language Models for Recipe Generation: A Comparative Analysis and Benchmark Study" \cite{vij2025finetuninglanguagemodelsrecipe} investigated the performance of smaller language models, including T5-small, SmolLM, and Phi-2, specifically for recipe generation tasks. The authors evaluated these models using both traditional NLP metrics (BLEU, ROUGE) and domain-specific metrics such as Ingredient Coverage and Step Complexity. While this study did not include larger models like GPT-4o or Llama3.1:8b that are examined in our work, it provides an important benchmark within the recipe domain that helps contextualize our findings. For broader model comparisons, established leaderboards such as the Hugging Face Open LLM Leaderboard\footnote{hf.com/open\_llm\_leaderboard} and LM Arena\footnote{https://lmarena.ai/} offer comprehensive evaluations across various benchmarks, which further support our experimental design and analysis framework.}

As LLMs have gained popularity, prompt engineering \cite{brown2020languagemodelsfewshotlearners} has emerged as a compelling alternative to fine-tuning approaches. Research on language models has focused on developing and evaluating prompt techniques across a diverse range of tasks. The research space has explored multiple prompting approaches: zero-shot learning \cite{promptingguide_zeroshot}, where models perform tasks without prior examples; few-shot learning \cite{promptingguide_fewshot}, which utilizes a small number of examples to guide performance, or more complex prompt techniques like MIPRO \cite{opsahlong2024optimizinginstructionsdemonstrationsmultistage}, which optimizes both instructions and few-shot demonstrations for multi-stage language model programs. The evolution of prompting techniques has demonstrated key advantages over traditional fine-tuning methods. 

Prompt engineering requires fewer computational resources and training data compared to full model fine-tuning. The well-built prompts can generate strong performance \cite{li2023overpromptenhancingchatgptefficient} across a range of tasks while maintaining model flexibility. These prompts have shown remarkable effectiveness across applications: in sentiment analysis, prompt-based fine-tuning achieved 92.7\% accuracy compared to traditional fine-tuning's 81.4\% while requiring only 32 examples on the SST-2 dataset \cite{gao2021makingpretrainedlanguagemodels}. Similarly, in classification tasks \cite{wei2023chainofthoughtpromptingelicitsreasoning} on datasets like SNLI, prompt-based methods with demonstrations achieved 79.7\% accuracy compared to standard fine-tuning's 48.4\%, demonstrating the approach's effectiveness across different linguistic tasks. Furthermore, prompting techniques can enhance the reasoning capabilities of language models. A notable example is the self-consistency approach introduced by Wang et al. \cite{wang2023selfconsistencyimproveschainthought}, which builds upon chain-of-thought prompting to improve complex reasoning tasks. This method leverages the intuition that multiple reasoning paths can lead to a correct answer, similar to human problem-solving processes. By sampling diverse reasoning paths instead of relying on a single greedy decoding path, their approach achieved striking improvements across reasoning benchmarks: GSM8K, SVAMP, and AQuA. 

These developments in prompting techniques, from basic few-shot demonstrations to more sophisticated approaches like self-consistency, highlight the potential of prompt engineering as a practical and effective method for enhancing language model performance. However, as these techniques become more sophisticated, the challenge of proper evaluation becomes increasingly important. The evaluation of LLMs in these contexts presents unique challenges, necessitating specialized metrics that assess both linguistic quality and compliance with domain-specific formatting requirements \cite{chang2024survey}. This has led to significant efforts in developing standardized evaluation methodologies, particularly for zero-shot text classification tasks. Notable contributions include Ribeiro et al.'s \cite{ribeiro2020accuracybehavioraltestingnlp} comprehensive framework for behavioral testing of natural language processing models.

The convergence of structured data approaches and LLM capabilities creates new opportunities for transforming unstructured text into standardized formats. While existing research demonstrates the potential of LLMs in domain-specific tasks and prompt engineering shows promise for enhancing model performance, there remains a critical need to systematically evaluate different LLM architectures and prompting techniques in the context of structured format generation. The following methodology addresses these requirements through a comprehensive evaluation framework that combines standard metrics like WER and ROUGE-L with specialized scoring mechanisms for assessing structured recipe generation to Cooklang format \cite{cooklang_2024}.


\added{While existing research demonstrates the significant progress in LLM applications for domain-specific tasks and the evolution of various prompting techniques, a critical gap remains in systematically evaluating LLMs' capability to transform unstructured text into standardized structured formats across different model architectures and prompting strategies. Prior studies have primarily focused on using LLMs for domain-specific knowledge extraction (as seen in food entity extraction research) or generation tasks (exemplified by RecipeNLG and LLaVA-Chef) but have not comprehensively assessed how different LLM architectures and prompting techniques affect the quality and accuracy of format conversion tasks. This gap is particularly significant as it represents the intersection of natural language understanding, structured data generation, and domain-specific knowledge representation—a capability with broad applications beyond culinary contexts to fields such as healthcare documentation, technical specification authoring, and legal document processing. Our work addresses this gap by providing a systematic comparative analysis of how different LLMs models sizes (from 8b to 70b parameters for open source) and prompting techniques (zero-shot, few-shot, and MIPROv2) influence performance in transforming unstructured recipe text into the standardized Cooklang format, offering insights applicable to structured text conversion challenges across domains. }

\section{Methodology}
\label{methodology}

\begin{figure*}
    \centering
    \includegraphics[width=1\linewidth]{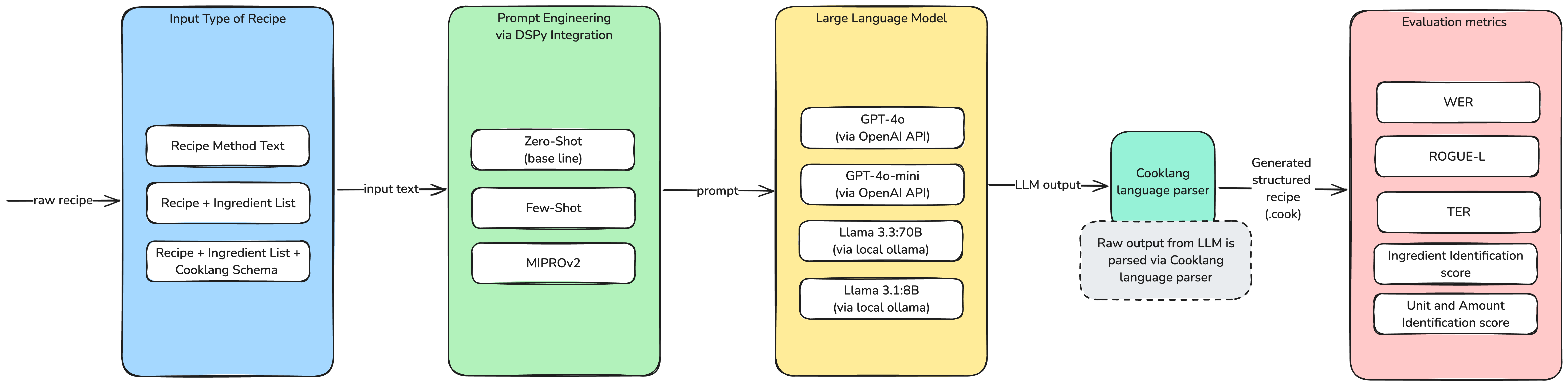}
    \caption{\added{Proposed methodology for evaluating the ability of Large Language Models in converting a recipe to Cooklang}}
    \label{fig:methodology_diagram}
\end{figure*}

This section contains the methodology for evaluating the effectiveness of Large Language Models (LLMs) in generating structured recipe specifications in Cooklang format \cite{cooklang_2024}. Our experimental design encompasses three key dimensions: input text variations, prompting strategies, and model architectures, allowing us to systematically assess how each factor influences the quality and accuracy of structured recipe generation. As illustrated in Figure \ref{fig:methodology_diagram}, our approach enables controlled comparison across these dimensions. We evaluated four state-of-the-art LLMs: Llama 3.1:8b, Llama 3.1:70b \cite{dubey2024llama3herdmodels}, \cite{meta_ai_2024}, \cite{llama_models_2024}, GPT-4o \cite{openai_gpt4o}, and GPT-4o-mini \cite{openai_gpt4o_mini}, each representing different points in the spectrum of model scale. These models were selected to provide an evaluation across both closed and open-source models, allowing us to compare performance between different model scales (from 8b to 70b parameters) and between proprietary and openly available solutions.

These LLMs represent the most popular models from both closed and open source, allowing for an assessment of performance across a range of capabilities. Through the testing of combinations of input formats, prompting strategies, and model architectures, the objective was to identify the most effective approach for converting recipe texts into the structured domain-specific recipe format. The recipe format contains two distinct input fields, as shown in dspy.Signatures (Listings \ref{lst:dspy_text2cooklang}, \ref{lst:dspy_text2cooklang_without_cooklang}, and \ref{lst:dspy_text2cooklang_without_cooklang_without_ings}). The first field contains ingredients for the recipe, presented as free text where each ingredient is comma-separated. The second field consists of the general recipe text, primarily, this field comprises cooking instructions. This field is converted to Cooklang format via LLM. The objective of this research is to examine the limitations of LLM in structured domain-specific text generation. The research explores how different models and prompt engineering influence semantic preservation and structural consistency in the conversion of unstructured recipes to formal specifications. Through an evaluation of model size and prompt engineering approaches, the research aims to advance our understanding of LLMs capabilities in maintaining semantic fidelity while adhering to strict syntactic constraints. This fundamental challenge in natural language processing extends far beyond recipe conversion to broader applications in structured knowledge representation.

 \added{Recipe formatting serves as an ideal benchmark for structured text conversion for several key reasons:
\begin{itemize}
    \item \textbf{Dual nature of content:} Recipes naturally combine narrative instructions with structured data elements (ingredients, measurements, cooking times), making them an excellent test case for evaluating both semantic understanding and structured data extraction.
    \item \textbf{Entity complexity:} The Cooklang format requires accurate identification of multiple entity types (ingredients, cookware, timers, comments) with their associated attributes (amounts, units), testing an LLM's fine-grained entity recognition capabilities.
    \item \textbf{Balancing precision and semantics:} Successful conversion requires both preserving the semantic meaning of instructions while adhering to strict syntactic rules defined by the Cooklang specification. 
    \item \textbf{Domain transferability: }The challenges in recipe formatting—identifying entities, preserving relationships, maintaining semantic equivalence—parallel those in other domains such as healthcare documentation, technical specifications, and legal documents. 
    \item \textbf{Clear evaluation criteria: }Recipe formatting provides objectively measurable outcomes (correct ingredient identification, proper syntax application) that allow for quantitative assessment of model performance. 
    \item \textbf{Accessibility:} Unlike highly specialized domains, recipes are universally understood, making the benchmark accessible while still maintaining sufficient complexity to challenge state-of-the-art models. 
\end{itemize}
}

\lstset{
    language=Python,
    basicstyle=\ttfamily\footnotesize,
    keywordstyle=\color{blue},
    stringstyle=\color{teal},
    commentstyle=\color{teal}\ttfamily,
    showstringspaces=false,
    breaklines=true,
    frame=single,
    caption={DSPy Program Class recipe text to Cooklang},
    label={lst:dspy_text2cooklang},
    morekeywords={dspy, dspy.InputField, dspy.OutputField, dspy.Signature}
}

\begin{lstlisting}
class CookLangSignature(dspy.Signature):
    """
    Convert plain recipe text with provided ingredients into Cooklang text format.
    Cooklang Recipe Specification:
        1. Ingredients
        - Use `@` to define ingredients
        - For multi-word ingredients, end with `{}`
        - Specify quantity in `{}` after the name
        - Use `%` to separate quantity and unit
        ```
        @salt
        @ground black pepper{}
        @potato{2}
        @bacon strips{1%kg}
        @syrup{1/2%tbsp}
        ```
        2. Comments
        - Single-line: Use `--` at the end of a line
        - Multi-line: Enclose in `[- -]`
        ```
        -- Don't burn the roux!
        Mash @potato{2%kg} until smooth -- alternatively, boil 'em first, then mash 'em, then stick 'em in a stew.
        ```
        3. Cookware
        - Define with `#`
        - Use `{}` for multi-word items
        ```
        #pot
        #potato masher{}
        ```
        4. Timers
        - Define with `~`
        - Specify duration in `{}`
        - Can include a name before the duration
        ```
        ~{25%minutes}
        ~eggs{3%minutes}
        ```
    Return only Cooklang formatted recipe, dont return any other information. Return whole recipe in Cooklang format! Dont stop till you reach the end of the recipe.
    """

    ingredients = dspy.InputField(desc="Ingredients for the recipe. Comma separated list of ingredients.")
    recipe_text = dspy.InputField(desc="Recipe text to convert to Cooklang format.")

    cooklang = dspy.OutputField(
        desc="Cooklang formatted recipe.",
    )

class CookLangFormatter(dspy.Module):
    def __init__(self):
        super().__init__()

        self.prog = dspy.ChainOfThought(CookLangSignature)

    def forward(self, recipe_text: str, ingredients: str) -> CookLangSignature:
        prediction = self.prog(recipe_text=recipe_text, ingredients=ingredients)
        return prediction
\end{lstlisting}

\lstset{
    language=Python,
    basicstyle=\ttfamily\footnotesize,
    keywordstyle=\color{blue},
    stringstyle=\color{teal},
    commentstyle=\color{teal}\ttfamily,
    showstringspaces=false,
    breaklines=true,
    frame=single,
    caption={DSPy Program Class recipe text to Cooklang without Cooklang Specification},
    label={lst:dspy_text2cooklang_without_cooklang},
    morekeywords={dspy, dspy.InputField, dspy.OutputField, dspy.Signature}
}

\begin{lstlisting}
class CookLangSignatureNoSteps(dspy.Signature):
    """
    Convert plain recipe text with provided ingredients into Cooklang text format.
    Return only Cooklang formatted recipe, dont return any other information. Return whole recipe in Cooklang format! Dont stop till you reach the end of the recipe.
    """

    ingredients = dspy.InputField(desc="Ingredients for the recipe. Comma separated list of ingredients.")
    recipe_text = dspy.InputField(desc="Recipe text to convert to Cooklang format.")

    cooklang = dspy.OutputField(
        desc="Cooklang formatted recipe.",
    )

class CookLangFormatterNoSteps(dspy.Module):
    def __init__(self):
        super().__init__()

        self.prog = dspy.ChainOfThought(CookLangSignatureNoSteps)

    def forward(self, recipe_text: str, ingredients: str) -> CookLangSignatureNoSteps:
        prediction = self.prog(recipe_text=recipe_text, ingredients=ingredients)
        return prediction
\end{lstlisting}

\lstset{
    language=Python,
    basicstyle=\ttfamily\footnotesize,
    keywordstyle=\color{blue},
    stringstyle=\color{teal},
    commentstyle=\color{teal}\ttfamily,
    showstringspaces=false,
    breaklines=true,
    frame=single,
    caption={DSPy Program Class recipe text to Cooklang without Cooklang Specification and without recipe ingredients},
    label={lst:dspy_text2cooklang_without_cooklang_without_ings},
    morekeywords={dspy, dspy.InputField, dspy.OutputField, dspy.Signature}
}

\begin{lstlisting}
class CookLangSignatureNoStepsNoIngredients(dspy.Signature):
    """
    Convert plain recipe text with provided ingredients into Cooklang text format.
    Return only Cooklang formatted recipe, dont return any other information. Return whole recipe in Cooklang format! Dont stop till you reach the end of the recipe.
    """

    recipe_text = dspy.InputField(desc="Recipe text to convert to Cooklang format.")

    cooklang = dspy.OutputField(
        desc="Cooklang formatted recipe.",
    )

class CookLangFormatterNoStepsNoIngredients(dspy.Module):
    def __init__(self):
        super().__init__()

        self.prog = dspy.ChainOfThought(CookLangSignatureNoStepsNoIngredients)

    def forward(self, recipe_text: str) -> CookLangSignatureNoStepsNoIngredients:
        prediction = self.prog(recipe_text=recipe_text)
        return prediction
\end{lstlisting}

The study implements three prompting strategies. The first strategy - Zero-Shot \cite{promptingguide_zeroshot} - established a baseline by presenting the task to the model without examples. The second strategy, Few-Shot \cite{reynolds2021prompt,promptingguide_fewshot}, involved the presentation of a limited number of examples to guide the model in its task. Final prompting strategy, implementation of MIPROv2 \cite{opsahlong2024optimizinginstructionsdemonstrationsmultistage}. The study tested three input configurations and ran every combination of input type and prompt technique across all four large language models. Following is an overview of the implementation of prompting strategies (method) and different variables at the input : 

\begin{enumerate}
    \item \textbf{Method}: The recipe method in its original format without additional processing.
    \item \textbf{Method + Ingredients}: The recipe method is accompanied by a list of recipe ingredients.
    \item \textbf{Method + Ingredients + Cooklang schema}: The recipe method and ingredients list, along with the Cooklang syntax specification.
\end{enumerate}

All configurations aimed to produce a string-modified version of recipe text, which we saved as a recipe.cook file and loaded it through the Cooklang parser. The parser loaded the file correctly, confirming that we had successfully converted the input text to Cooklang format. To fully evaluate our approach, we tested all possible combinations of input types, prompt techniques, and LLMs.

\subsection{Prompt engineering}

We based our approach on transforming recipes into Cooklang format using an LLM on the DSPy framework \cite{khattab2023dspy}. We developed distinct DSPy programs for each model input type to ensure optimal performance across all scenarios. We implemented few-shot prompting using the bootstrap-few-shot-with-random-search method from the DSPy library, keeping the default hyperparameters. Similarly, we implemented MIPROv2 \cite{opsahlong2024optimizinginstructionsdemonstrationsmultistage} using the existing DSPy implementation with its default hyperparameters. The Listings \ref{lst:dspy_text2cooklang}, \ref{lst:dspy_text2cooklang_without_cooklang} show the DSPy programs we used in our implementation. These prompts form the core elements of our recipe transformation process. \added{Examples of prompts for Few-Shot and MIPROv2 could be found in Listings \ref{fig:cooklang-prompt-1} and \ref{fig:cooklang-prompt-2}, respectively. }

\lstdefinestyle{promptstyle}{
    backgroundcolor=\color{white},
    basicstyle=\footnotesize\ttfamily,
    breakatwhitespace=false,
    breaklines=true,
    captionpos=b,
    commentstyle=\footnotesize\ttfamily, 
    keywordstyle=\footnotesize\ttfamily, 
    stringstyle=\footnotesize\ttfamily,  
    identifierstyle=\footnotesize\ttfamily, 
    keepspaces=true,
    numbers=left,
    numbersep=5pt,
    numberstyle=\tiny,
    showspaces=false,
    showstringspaces=false,
    showtabs=false,
    tabsize=2,
    frame=single,
    framesep=5pt,
    framerule=0pt,
    language={},
    alsoletter={@\#\%},
}

\begin{figure*}[htbp]
    \centering
\begin{lstlisting}[
        style=promptstyle,
        caption={\added{Example of prompt (Few-Shot) for Cooklang conversion without cooklang notation and with ingredients}},
        label={fig:cooklang-prompt-1}
    ]
Convert plain recipe text with provided ingredients into Cooklang text format.
Return only Cooklang formatted recipe, dont return any other information. Return whole recipe in Cooklang format! Dont stop till you reach the end of the recipe.

---

Ingredients: 50 g ciabatta,some oil,,50 g cottage cheese,80 g smoked salmon,5 g capers,1 g dill,some lemon,some black pepper
Recipe Text: Slice the whole ciabatta loaf in half and then slice the halves lengthwise. Fry the ciabatta halves in a frying pan in a little oil, a griddle pan gives the best results and makes nice lines as the bread toasts. Spread the cottage cheese on the toasted ciabatta liberally, cut out the smoked salmon slices and place on the bread with a sprinkling of capers and finally chopped dill. Serve with a lemon wedge and freshly cracked black pepper.
Cooklang: Slice the whole @ciabatta{50%g} loaf in half and then slice the halves lengthwise. Fry the ciabatta halves in a #frying pan{} in a little @oil, a griddle pan gives the best results and makes nice lines as the bread toasts. Spread the @cottage cheese{50%g} on the toasted ciabatta liberally, cut out the @smoked salmon{80%g} slices and place on the bread with a sprinkling of @capers{5%g} and finally chopped @dill{1%g}. Serve with a @lemon wedge and freshly cracked @black pepper{}.

[Additional examples omitted for brevity]

---

Follow the following format.

Ingredients: Ingredients for the recipe. Comma separated list of ingredients.

Recipe Text: Recipe text to convert to Cooklang format.

Reasoning: Let's think step by step in order to ${produce the cooklang}. We ...

Cooklang: Cooklang formatted recipe.

---

Ingredients: {ingredients}

Recipe Text: {recipe text}

Reasoning: Let's think step by step in order to Cooklang:
\end{lstlisting}
\end{figure*}

\begin{figure*}[htbp]
    \centering
\begin{lstlisting}[
        style=promptstyle,
        caption={\added{Example of prompt (MIPROv2) for Cooklang conversion with cooklang notation and with ingredients}},
        label={fig:cooklang-prompt-2}
    ]
Transform the given plain text recipe and accompanying ingredients into a structured Cooklang format. Use the following specifications:

1. **Ingredients:**
   - Prefix each ingredient with `@`.
   - For multi-word ingredients, wrap the entire ingredient in `{}`.
   - Indicate the quantity directly after the ingredient name in `{}`, followed by a `%` for the unit.
   - Example: `@salt`, `@ground black pepper{}`, `@potato{2}`, `@bacon strips{1%kg}`, `@syrup{1/2%tbsp}`.

2. **Comments:**
   - Use `--` to denote single-line comments.
   - For multi-line comments, enclose them in `[- -]`.
   - Example: `-- Don't burn the roux!`.

3. **Cookware:**
   - Use `#` to specify the cookware required.
   - For multi-word items, enclose in `{}`.
   - Example: `#pot`, `#potato masher{}`.

4. **Timers:**
   - Use `~` to specify timing instructions.
   - Indicate the duration in `{}`.
   - You may include a name before the duration.
   - Example: `~{25%minutes}`, `~eggs{3%minutes}`.

Return only the Cooklang formatted recipe without any additional commentary or information. Ensure that the output is complete and captures all elements of the recipe succinctly.

---

[Few-shot examples omitted for brevity]

---

Follow the following format.

Ingredients: Ingredients for the recipe. Comma-separated list of ingredients.

Recipe Text: Recipe text to convert to Cooklang format.

Reasoning: Let's think step by step in order to ${produce the cooklang}. We ...

Cooklang: Cooklang formatted recipe.

---

Ingredients: {ingredients}

Recipe Text: {recipe text}

Reasoning: Let's think step by step in order to
\end{lstlisting}
\end{figure*}

\subsection{Dataset}

We assembled the dataset for this research from sources that provide recipes in both standard and Cooklang formats. The primary sources include the Cooklang documentation and associated GitHub repositories \cite{cooklang_recipes,cooklang_spec_examples,awesome_cooklang_recipes}. We merged the recipes from these sources to create a comprehensive dataset, which we now make publicly available in our GitHub  repository\footnote{\url{https://github.com/williambrach/llm-text2cooklang/tree/main/data}}. The dataset contains \replaced{32}{1098} \deleted{distinct} recipe samples, representing a \deleted{diverse} range of recipe categories, including baking, breakfast, dinners, lunches, and soups. This diversity ensures a broad representation of culinary styles and techniques, enabling a more robust evaluation of the models. We enhanced the dataset by extracting ingredients from the Cooklang Specification using the Cooklang CLI \cite{cooklang_github_cli}. We added these extracted ingredients as an additional feature to the dataset, enriching the information available for analysis and prompt optimization.

\subsection{Evaluation Process}

We evaluated the models across several metrics, including overall performance, the impact of different prompt techniques, the effect of using Cooklang and ingredient exclusion, and domain-specific tasks related to recipe understanding. We tested the LLMs ability to accurately identify ingredients, units, and amounts. We used several metrics to assess the LLMs performance in detail.
\begin{enumerate}
    \item \textbf{Word Error Rate (WER)} \cite{wer_ali2018word,wer_klakow2002testing}: The metric measures the edit distance between generated and reference text quantitatively, normalizing values by the reference text's word length. A lower WER value indicates more accurate performance that comes closer to the desired outcome.
    \item \textbf{ROUGE-L} \cite{google2019rouge,google_see2017pointsummarizationpointergeneratornetworks}: The metric to evaluate the quality of generated summaries or translations. ROUGE-L calculates F-scores based on the length of the longest common subsequence between the candidate and reference texts. This method allows for capturing sentence-level structure similarity and identifying the longest co-occurring in-sequence n-grams automatically. Higher ROUGE-L scores indicate better performance, suggesting that the generated text shares more and longer sequential word matches with the reference text.
    \item \textbf{Token Error Rate (TER)}: This metric measures how closely generated Cooklang texts match their references by calculating a normalized edit distance. The system first tokenizes texts by breaking them down into Cooklang-specific elements (ingredients, cookware, and cooking steps). It then computes TER for each element by counting the minimum number of edits needed to transform the generated text into the reference - including insertions, deletions, substitutions, and shifts of Cooklang tokens - and divides this count by the total tokens in the reference. Lower TER scores show better performance.    
    \item \textbf{Ingredient Identification Score}: Metric assesses the LLM capacity to accurately identify all ingredients in a given recipe. A score of 1 indicates that the LLM successfully identified all ingredients present in the reference recipe, whereas a score of 0 indicates that the LLM failed to identify all ingredients correctly.
    \item \textbf{Unit and Amount Identification Scores}: These two scores are metrics that assess the LLMs accuracy in identifying units of measurement and ingredient amounts in a recipe. For each metric, the score is 1 if all units or amounts are correctly identified for every ingredient in the recipe and 0 if one unit or amount is incorrectly identified or missing. 
\end{enumerate}

For each configuration, we evaluated all metrics across the test samples to obtain individual scores. The mean value for each metric $M$ under configuration $c$ was calculated as:

$$\bar{M_c} = \frac{1}{N} \sum_{i=1}^{N} M_{c,i}$$

where $M_{c,i}$ represents the metric score for configuration $c$ and test sample $i$, and $N$ denotes the total number of test samples. This approach provides a comprehensive view of LLM performance across all aspects of recipe understanding and generation tasks. Analyzing these aggregated results helps us identify strengths and weaknesses in different model-prompt configurations and determine the most effective approaches for recipe-related natural language processing.

\subsection{Hardware and Deployment Setup}
We deployed all open-source models (Llama3.1:8b and \replaced{LLama3.1:70b}{LLama3.3:70b}) locally using ollama \cite{ollama2024}, a framework that runs and manages local deployment of LLMs. Our local setup included two NVIDIA GeForce RTX 4090 graphics cards, which provided the necessary computational power for efficient model inference. We accessed OpenAI GPT models through the OpenAI API.

\subsection{Computational Overhead}

\added{Our methodology's computational requirements can be analyzed across three phases:}

\added{\begin{itemize}
    \item \textbf{Training:} The DSPy framework introduces upfront computational overhead during training as it runs multiple prompt variations to select optimal configurations. This is a one-time cost; once optimal programs are identified, they can be reused during deployment.
    \item \textbf{Inference:} The computational overhead during inference is negligible, primarily affecting the number of input tokens processed by the LLM. Since input token processing is less computationally expensive than output token generation, this represents a minor overhead compared to base LLM inference costs.
    \item \textbf{Implementation:} The DSPy framework has minimal implementation overhead, efficiently constructing prompts without introducing significant processing loops or memory constraints. The memory footprint is minimal, consisting primarily of Python objects composed of strings.
\end{itemize}}

\added{In summary, our approach introduces a one-time training cost but maintains efficient performance during deployment with minimal additional computational requirements compared to standard LLM operation.}

\section{Results}
\label{results}

This section analyzes the performance of four Large Language Models (LLMs): GPT-4o, GPT-4o-mini,\replaced{LLama3.1:70b}{LLama3.3:70b}, and Llama3.1:8b. The analysis evaluates each model using three traditional metrics: Word Error Rate (WER), ROUGE-L score, and Translation Edit Rate (TER), grouping all results by their mean values. Table \ref{tab:model_performance_traditional} shows each model's performance metrics:

\begin{table}[htbp]
\centering
\caption{Mean WER, ROUGE-L, and TER Scores Across Language Models}
\label{tab:model_performance_traditional}
\resizebox{\columnwidth}{!}{%
\begin{tabular}{@{}lcccc@{}}
\toprule
\textbf{Model} & \textbf{WER} ↓ & \textbf{ROUGE-L} ↑ & \textbf{TER} ↓ \\
\midrule
\textbf{GPT-4o} & \replaced{\textbf{0.1996}}{\textbf{0.3893}} & \replaced{\textbf{0.9280}}{\textbf{0.8128}} & \replaced{\textbf{0.8619}}{\textbf{1.5255}}\\
GPT-4o-mini & \replaced{0.2907}{0.4808} & \replaced{0.8689}{0.7550} & \replaced{1.4088}{2.0381}\\
\replaced{Llama3.1:70b}{Llama3.3:70b} & \replaced{0.9803}{0.5510} & \replaced{0.5453}{0.7382} & \replaced{4.7859}{2.2802}\\
Llama3.1:8b & \replaced{1.3513}{0.5921} & \replaced{0.4526}{0.6911} & \replaced{6.4910}{2.6088}\\
\bottomrule
\end{tabular}%
}
\caption*{\small Note: $\downarrow$ indicates lower values are better, $\uparrow$ indicates higher values are better.}
\end{table}

\begin{figure*}[htbp]
    \centering
    \includegraphics[width=\linewidth]{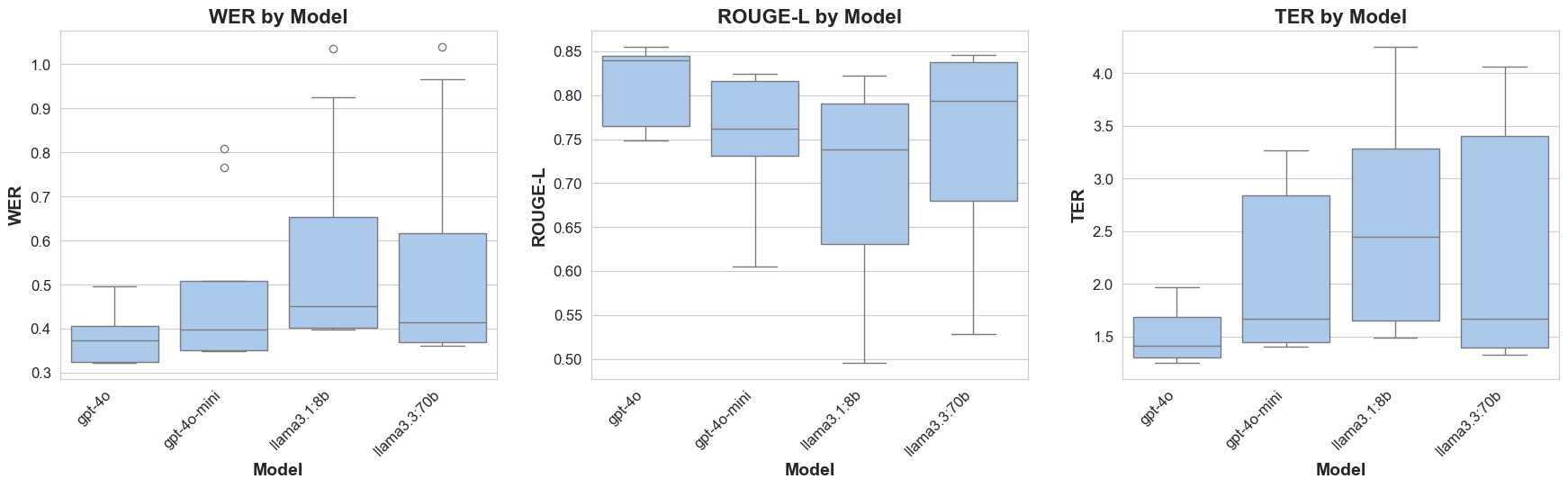}
    \caption{\added{Comparison of Language Model Performance Across WER, ROUGE-L, and TER Metrics}}
    \label{fig:model_comparison_standard}
\end{figure*}

As shown in Figure \ref{fig:model_comparison_standard}, the GPT-4o model displays superior performance across all metrics in converting inputs to the Cooklang format. The model achieved the highest ROUGE-L score (\replaced{0.9280}{0.8128}), indicating an overlap between the generated Cooklang output and the reference text at the n-gram level. Furthermore, the model had the lowest Token Error Rate (TER) of \replaced{0.8619}{1.5255}, indicating minimal hallucination or generation of content not grounded in the original text. Additionally, it showed the lowest Word Error Rate (WER) of \replaced{0.1996}{0.3893}, implying high accuracy in maintaining the correct word sequence with minimal insertions, substitutions, or deletions when converting to Cooklang format. These results suggest that GPT-4o could be particularly well-suited for applications requiring accurate transformation between different text formats and structures, as its strong performance in maintaining semantic content while adapting syntactic structure indicates robust capabilities in text transformation tasks. GPT-4o-mini ranks second in converting plain text to Cooklang, demonstrating consistent performance, but falls behind GPT-4o. Its Word Error Rate (WER) of \replaced{0.2907}{0.4808} shows more errors in word sequence maintenance. The lower ROUGE-L score of \replaced{0.8689}{0.7550} reveals reduced overlap between the generated Cooklang and reference text. A Token Error Rate (TER) of \replaced{1.4088}{2.0381} indicates that GPT-4o-mini generates more content without grounding in the original text. Results for the family of Llama3.1 demonstrate that models exhibit considerably reduced performance in Cooklang conversion relative to the GPT models. The 70b version exhibits superior performance compared to the 8b version across all metrics, suggesting that an increased model size contributes to enhanced performance. However, both models demonstrate a notable deficit in comparison to the GPT models. The WER values for the Llama models exceed \replaced{0.98}{0.55}, indicating a significant prevalence of errors in word sequence. The ROUGE-L scores are below \replaced{0.55}{0.73}, suggesting a \replaced{notable discrepancy between }{similar performance as GPT-4o-mini in} the generated text and the reference text. Additionally, the TER values exceed \replaced{4.78}{2.28}, indicating a \replaced{considerable}{possible} tendency for hallucination or the generation of content that is not directly relevant to the source text. 

\subsection{Impact of Prompt Techniques on Model Performance}
\begin{figure*}
    \centering
    \includegraphics[width=\linewidth]{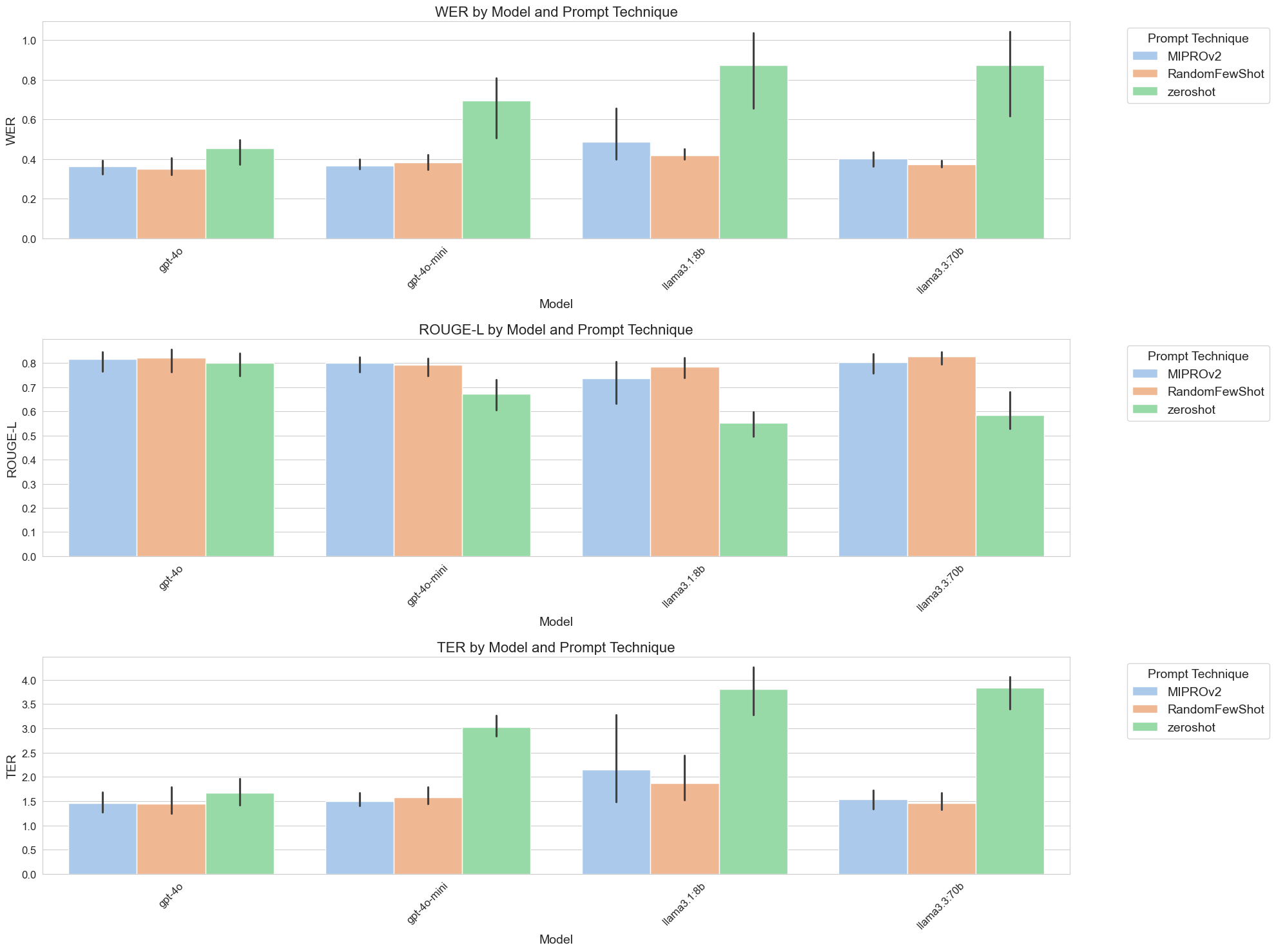}
    \caption{\added{Prompting Technique Performance Across WER, ROUGE-L, and TER Metrics}}
    \label{fig:model_comparison_standard_by_prompt}
\end{figure*}

Analysis of the layer of prompt techniques -- MIPROv2, Few-Shot, Zero-Shot -- reveals differences in their effectiveness across models. The results, presented in Table \ref{tab:model_performance_prompt} and Figure \ref{fig:model_comparison_standard_by_prompt}, demonstrate that the impact of prompt techniques varies considerably depending on the model used. The Few-Shot prompt technique consistently demonstrates superior performance compared to other techniques across all models, achieving the lowest word error rate (WER), the highest ROUGE-L score, and the lowest term error rate (TER). 

The efficacy of each prompt technique varies considerably across different models. Larger, more advanced models (GPT-4o and GPT-4o-mini) demonstrate superior overall performance and less variability between prompt techniques in comparison to smaller models (\replaced{LLama3.1:70b}{LLama3.3:70b} and Llama3.1:8b). It is noteworthy that while Few-Shot consistently ranks first, \replaced{the relative performance of MIPROv2 and Zero-Shot varies depending on the model and metric.}{but MIPROv2 optimized prompts are closely following. Zero-shot prompts show a consistent gap to other techniques for each model.} Larger models demonstrate the most optimal overall performance across all techniques, with Few-Shot attaining a WER as low as \replaced{0.073}{0.3509} for GPT-4o. In contrast, Llama3 3.1 models exhibit elevated error rates and \replaced{diminished}{lower} ROUGE-L scores.
\deleted{with Zero-Shot outperforming MIPROv2, particularly for the 8b variant.}

\begin{table}[htbp]
\centering
\caption{Model Performance Metrics Across Prompt Techniques}
\label{tab:model_performance_prompt}
\resizebox{\columnwidth}{!}{%
\begin{tabular}{@{}lccc@{}}
\toprule
\textbf{Model} & \textbf{WER} ↓ & \textbf{ROUGE-L} ↑ & \textbf{TER} ↓ \\
\midrule
GPT-4o (MIPROv2) & \replaced{0.2221}{0.3637} & \replaced{0.9384}{0.8171} & \replaced{0.9950}{1.4566} \\
\textbf{GPT-4o (Few-Shot)} & \replaced{\textbf{0.0730}}{\textbf{0.3509}} & \replaced{\textbf{0.9722}}{0.8209} & \replaced{\textbf{0.2096}}{\textbf{1.4467}} \\
GPT-4o (Zero-Shot) & \replaced{0.3038}{0.4533} & \replaced{0.8733}{0.8005} & \replaced{1.3811}{1.6733} \\
GPT-4o-mini (MIPROv2) & \replaced{0.1962}{0.3664} & \replaced{0.8948}{0.8008} & \replaced{1.1047}{1.5024} \\
GPT-4o-mini (Few-Shot) & \replaced{0.0865}{0.3819} & \replaced{0.9646}{0.7928} & \replaced{0.2509}{1.5820} \\
GPT-4o-mini (Zero-Shot) & \replaced{0.6208}{0.6943} & \replaced{0.7387}{0.6715} & \replaced{2.9723}{3.0299} \\
\replaced{Llama3.1:70b}{Llama3.3:70b} (MIPROv2) & \replaced{1.2060}{0.4037} & \replaced{0.4100}{0.8023} & \replaced{5.9615}{1.5364} \\
\replaced{Llama3.1:70b}{Llama3.3:70b} (Few-Shot) & \replaced{0.8692}{0.3747} & \replaced{0.6170}{\textbf{0.8270}} & \replaced{4.4031}{1.4633} \\
\replaced{Llama3.1:70b}{Llama3.3:70b} (Zero-Shot) & \replaced{0.8656}{0.8746} & \replaced{0.6091}{0.5854} & \replaced{3.9931}{3.8408} \\
Llama3.1:8b (MIPROv2) & \replaced{1.9504}{0.4869} & \replaced{0.2806}{0.7371} & \replaced{9.2570}{2.1442} \\
Llama3.1:8b (Few-Shot) & \replaced{1.1835}{0.4172} & \replaced{0.4898}{0.7837} & \replaced{5.8003}{1.8714} \\
Llama3.1:8b (Zero-Shot) & \replaced{0.9200}{0.8721} & \replaced{0.5874}{0.5526} & \replaced{4.4156}{3.8108} \\
\bottomrule
\end{tabular}%
}
\caption*{\small Note: $\downarrow$ indicates lower values are better, $\uparrow$ indicates higher values are better.}
\end{table}

\subsection{Effect of Cooking Language Specification and Ingredient Exclusion}

We also evaluated whether inserting Cooklang language specifications into prompts or presenting a list of ingredients could yield a significant difference. Our findings indicate considerable discrepancies in model performance contingent on the utilization of Cooklang and the exclusion of ingredient information. Examined four models under different configurations: with/without Cooklang specification and with/without recipe ingredients. The results, presented in Table \ref{tab:model_performance_with_without}, demonstrate that these factors influence model performance across evaluation metrics.

\begin{table}[htbp]
\centering
\caption{LLM Performance Metrics Across Different Configurations}
\label{tab:model_performance_with_without}
\resizebox{\columnwidth}{!}{%
\begin{tabular}{@{}lcccccc@{}}
\toprule
\textbf{Model} & \textbf{Cooklang Specification} & \textbf{Ingredients} &  \textbf{WER} ↓ & \textbf{ROUGE-L} ↑ & \textbf{TER} ↓ \\
\midrule														
GPT-4o & False & False & \replaced{0.3370}{0.4292} & \replaced{0.8839}{0.7585} & \replaced{1.4448}{1.8141} \\
GPT-4o & False & True & \replaced{0.1596}{0.3825} & \replaced{0.9392}{0.8350} & \replaced{0.6676}{1.4019} \\
\textbf{GPT-4o} & \textbf{True} & \textbf{True} & \replaced{\textbf{0.1023}}{\textbf{0.3563}} & \replaced{\textbf{0.9608}}{\textbf{0.8449}} & \replaced{\textbf{0.4733}}{\textbf{1.3605}} \\
GPT-4o-mini & False & False & \replaced{0.3924}{0.5429} & \replaced{0.8461}{0.7053} & \replaced{1.5210}{2.242} \\
GPT-4o-mini & False & True & \replaced{0.2768}{0.4970} & \replaced{0.8800}{0.7680} & \replaced{1.2981}{1.9757} \\
GPT-4o-mini & True & True & \replaced{0.2075}{0.4026} & \replaced{0.8770}{0.7919} & \replaced{1.4443}{1.8965} \\
\replaced{Llama3.1:70b}{Llama3.3:70b} & False & False & \replaced{1.1409}{0.5972} & \replaced{0.4483}{0.6937} & \replaced{5.3866}{2.4857} \\
\replaced{Llama3.1:70b}{Llama3.3:70b} & False & True & \replaced{1.2036}{0.6058} & \replaced{0.4926}{0.7329} & \replaced{5.8035}{2.3081} \\
\replaced{Llama3.1:70b}{Llama3.3:70b} & True & True & \replaced{0.5964}{0.4501} & \replaced{0.6951}{0.7880} & \replaced{3.1677}{2.0466} \\
Llama3.1:8b & False & False & \replaced{1.6832}{0.5766} & \replaced{0.3529}{0.6863} & \replaced{8.1012}{2.4011} \\
Llama3.1:8b & False & True & \replaced{1.2686}{0.6127} & \replaced{0.4796}{0.7201} & \replaced{5.7104}{2.4220} \\
Llama3.1:8b & True & True & \replaced{1.1021}{0.5869} & \replaced{0.5252}{0.6669} & \replaced{5.6614}{3.0033} \\
\bottomrule
\end{tabular}%
}
\caption*{\small Note: $\downarrow$ indicates lower values are better, $\uparrow$ indicates higher values are better.}
\end{table}

\begin{figure*}
    \centering
    \includegraphics[width=\linewidth]{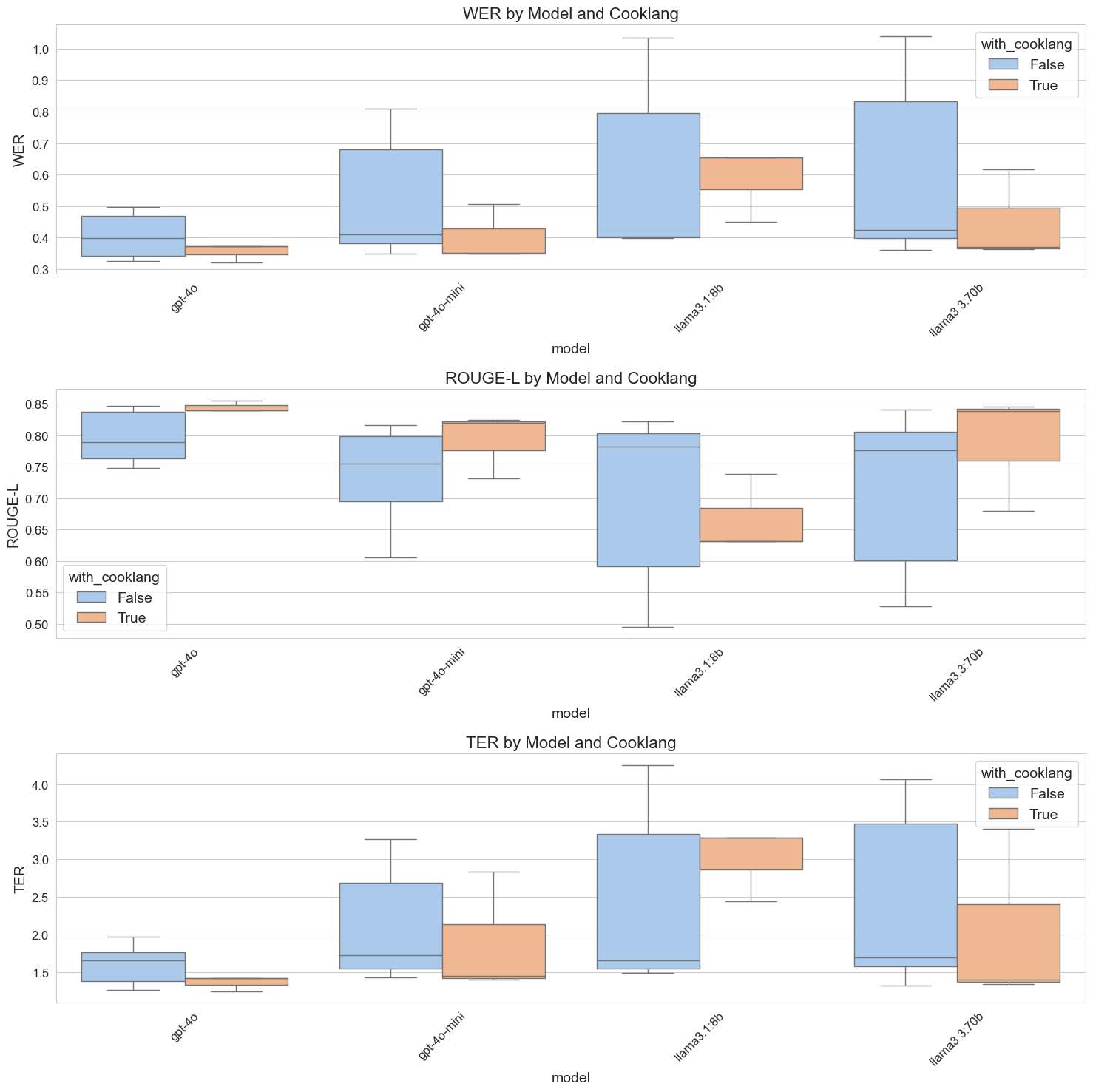}
    \caption{\added{Impact of Cooklang Specification Integration on WER, ROUGE-L, and TER Performance Metrics}}
    \label{fig:model_comparison_standard_by_cooklang}
\end{figure*}

\begin{figure*}
    \centering
    \includegraphics[width=\linewidth]{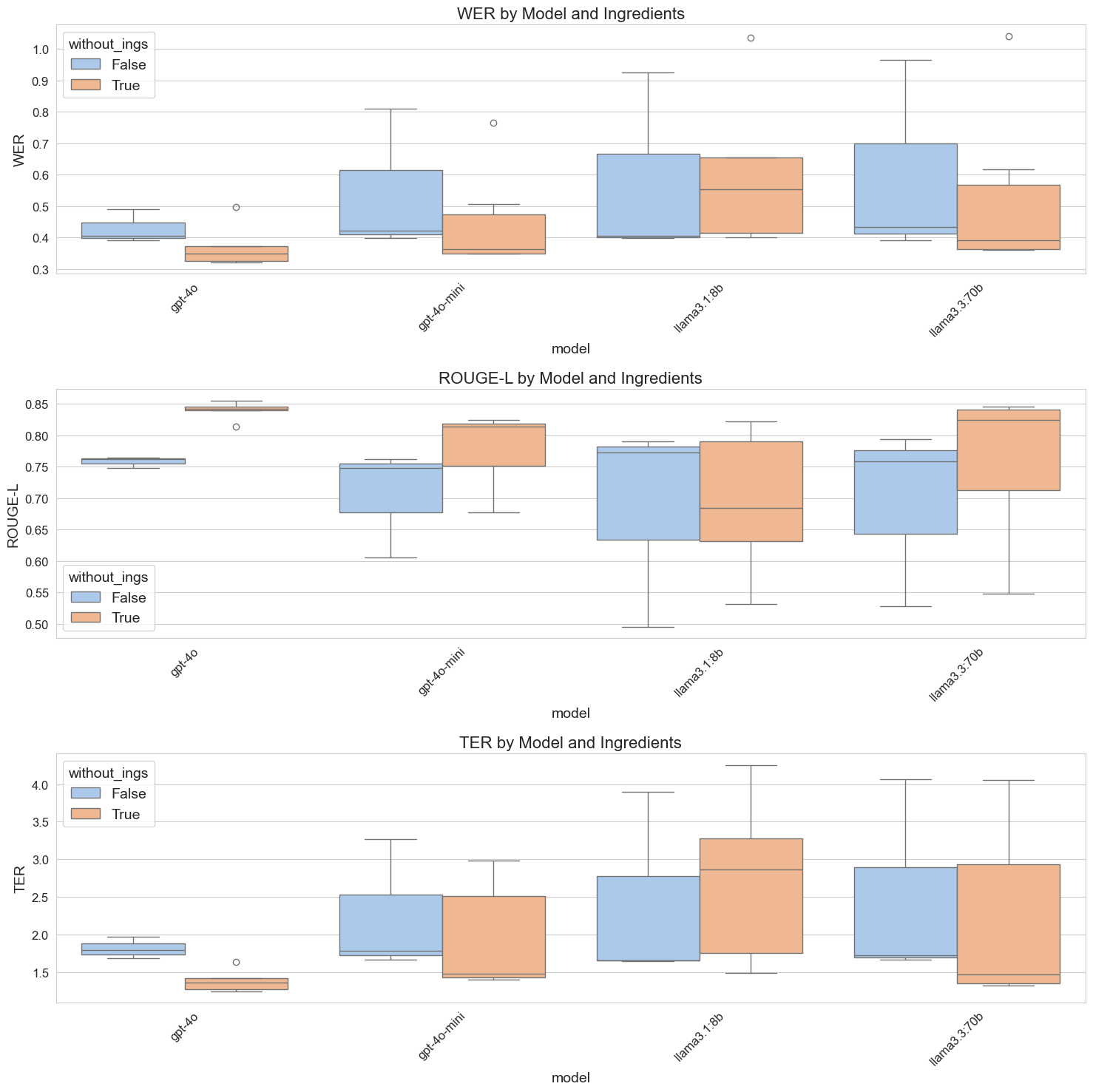}
    \caption{\added{Impact of Ingredients Integration on WER, ROUGE-L, and TER Performance Metrics}}
    \label{fig:model_comparison_standard_by_ings}
\end{figure*}

Figures \ref{fig:model_comparison_standard_by_cooklang} and \ref{fig:model_comparison_standard_by_ings} show that including both Cooklang and ingredient information improved performance across all tested models. This suggests that structured language and contextual information enhance recipe processing capabilities. GPT-4o performed best across all configurations and metrics, achieving its highest scores when using both Cooklang and ingredient information. The benefits of these additions scaled with model size, as larger models showed more substantial improvements. While ingredient information alone generally improved performance, adding Cooklang created further gains, highlighting the value of domain-specific structured languages. These findings could indicate that language models benefit from both structured formats and comprehensive contextual information when processing domain-specific tasks. The pronounced impact of including ingredient information, even without Cooklang specifications, indicates that contextual knowledge plays a crucial role in recipe understanding. This aligns with previous findings in natural language processing, where domain-specific context enhances task performance. However, the synergistic effect of combining both Cooklang and ingredients suggests that structured formats help models better leverage available contextual information.

\subsection{Ingredient Identification}
\begin{figure}
    \centering
    \includegraphics[width=\columnwidth]{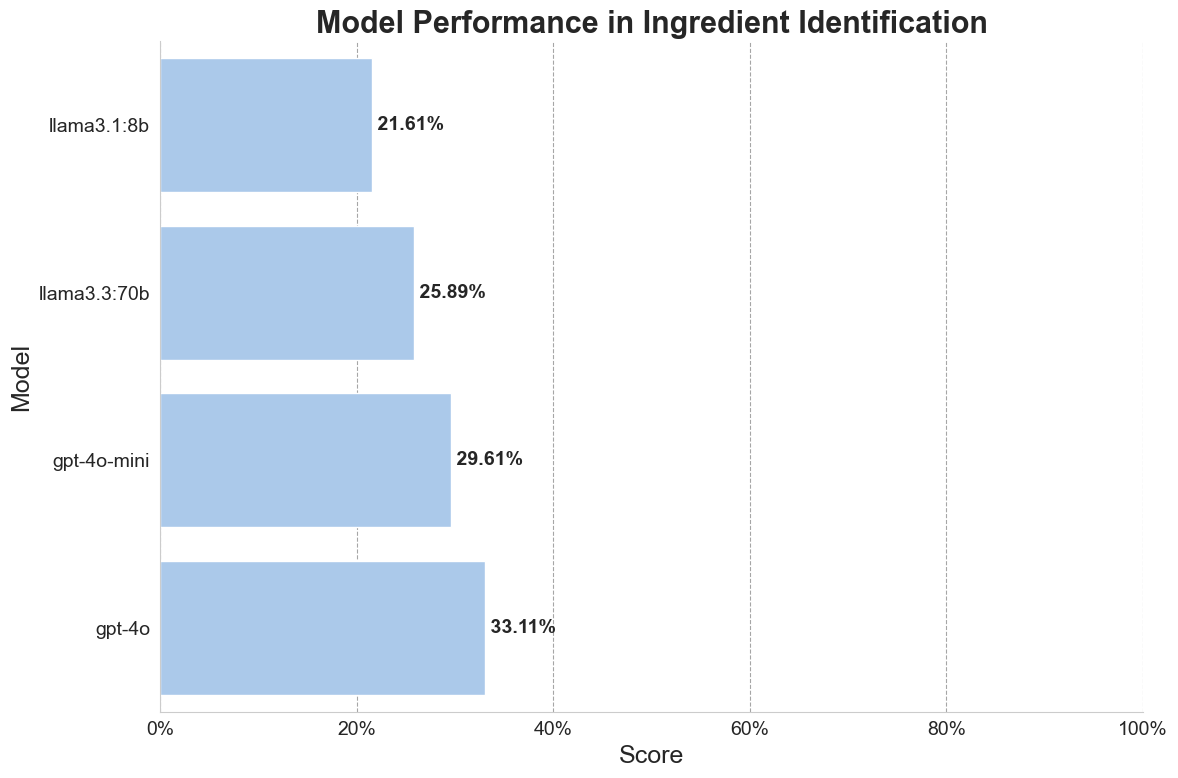}
    \caption{\added{Model Performance in Ingredient Identification, higher values are better.}}
    \label{fig:ing_Identification}
\end{figure}

\begin{figure*}
    \centering
    \includegraphics[width=0.9\linewidth]{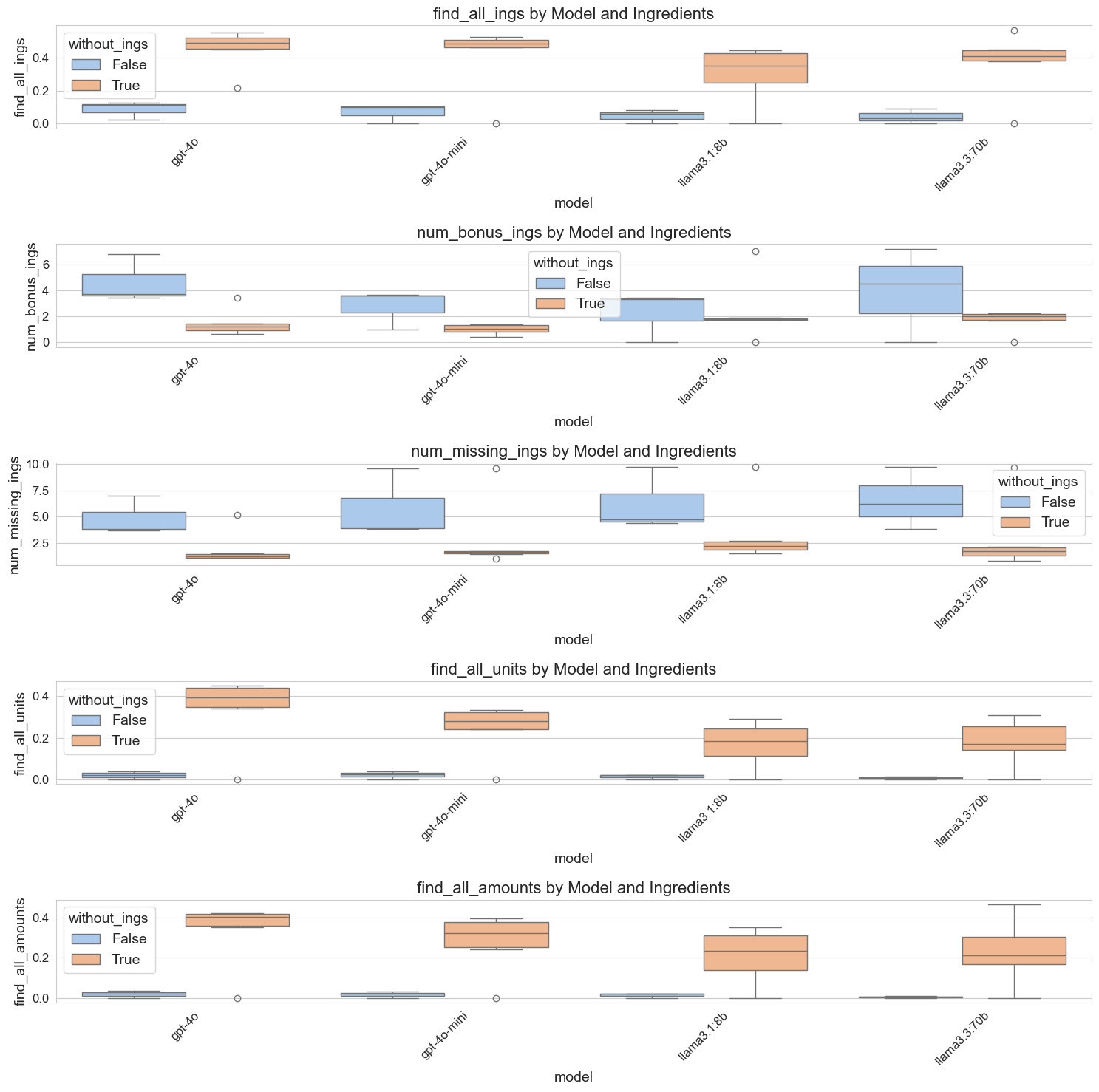}
    \caption{\added{Comparison of inputting ingredients into model prompt}}
    \label{fig:model_comparison_standard_by_model_ings}
\end{figure*}

Our evaluation metric assessed how accurately LLMs identify all ingredients in recipes. Figure \ref{fig:ing_Identification} illustrates the identification capabilities and shows performance variations across four tested models. GPT-4o achieved the highest score of \replaced{0.61}{0.33} on the ingredient identification metric (0-1 range), while GPT-4o-mini scored \replaced{0.56}{0.296}. \added{The larger LLama3.3:70b closely followed GPT-4o-mini with 0.2589 (difference 0.0371)}. \replaced{The LLama models }{The smaller LLama3.1:8b} performed \deleted{significantly} worse \deleted{with LLama3.3:70b scoring 0.24} and \deleted{Llama3.1:8b} reaching only 0.216. This gap reveals a \replaced{significant}{notable} difference between GPT and Llama models in ingredient identification tasks. \deleted{The top GPT model outperforms the Llama model by approximately 2.5 times. Within the Llama family, increasing the model size from 8b to 70b parameters improved performance by 2.4 times, highlighting how the model scale affects this task.} It is important to note that this evaluation methodology relies solely on exact string matching for ingredient identification without accounting for spelling variations or minor textual discrepancies. This strict matching criterion likely underestimates the models' true ingredient identification capabilities, as it penalizes semantically correct but textually inexact matches. Future work could enhance this metric by incorporating fuzzy matching or semantic similarity measures to provide a more comprehensive assessment of ingredient identification performance. Figure \ref{fig:model_comparison_standard_by_model_ings} provides a deeper analysis, examining both missing ingredients (false negatives) and incorrectly added ingredients (false positives) across models.

\subsection{Accuracy in Unit and Amount Identification}
Table \ref{tab:acc_units_and_amounts} shows LLM performance metrics across configurations and prompt techniques, focusing on unit and amount identification accuracy in domain-specific contexts. We evaluated the models using MIPROv2, few-shot, and zero-shot prompting techniques. GPT-4o, with \replaced{few-shot prompting}{MIPROv2 optimized prompt}, achieved the highest scores, reaching \replaced{0.5729 and 0.5833}{0.3017 and 0.2900} for unit and amount identification. Few-Shot and MIPROv2 techniques outperformed Zero-Shot across most models, highlighting the impact of effective prompting. GPT-4o and GPT-4o-mini surpassed the Llama models in performance, while the 70b Llama version exceeded its 8b variant \added{but, this metric was the hardest one for LLama3.3:70 with the largest gap it scored against GPT-4o-mini}. Zero-shot yielded the lowest results, demonstrating the necessity of proper prompting. Llama models, particularly the 8b version, struggled compared to GPT models. Notably, Llama3.1:8b showed slightly better performance with Zero-Shot for amount detection. These results demonstrate how model size and prompting techniques affect specialized task performance while indicating room for improvement in domain applications. Similar to ingredient identification, our metrics that evaluate LLMs across Unit and Amount Identification rely solely on exact string matching for ingredient identification without accounting for spelling variations or minor textual discrepancies.

\begin{table}[htbp]
\centering
\caption{LLM Performance Metrics Across Different Models and Prompt Techniques}
\label{tab:acc_units_and_amounts}
\resizebox{\columnwidth}{!}{%
\begin{tabular}{@{}lcccc@{}}
\toprule
\textbf{Model} & \textbf{Prompt Technique} & \textbf{Find All Units} ↑ & \textbf{Find All Amounts} ↑ \\
\midrule
				
GPT-4o & MIPROv2 & \replaced{0.5313}{\textbf{0.3017}} & \replaced{0.5625}{\textbf{0.29}} \\
\textbf{GPT-4o} & \textbf{Few-Shot} & \replaced{\textbf{0.5729}}{0.2433} & \replaced{\textbf{0.5833}}{0.2533} \\
GPT-4o & Zero-Shot & \replaced{0.2292}{0.15} & \replaced{0.2396}{0.14} \\
GPT-4o-mini & MIPROv2 & \replaced{0.5234}{0.2033} & \replaced{0.5469}{0.2050} \\
GPT-4o-mini & Few-Shot & \replaced{0.4688}{0.2250} & \replaced{0.4479}{0.2350} \\
GPT-4o-mini & Zero-Shot & \replaced{0.1146}{0.0800} & \replaced{0.1563}{0.1317} \\
\replaced{Llama3.1:70b}{Llama3.3:70b} & MIPROv2 & \replaced{0.1771}{0.0983} & \replaced{0.1667}{0.1183} \\
\replaced{Llama3.1:70b}{Llama3.3:70b} & Few-Shot & \replaced{0.1979}{0.1600} & \replaced{0.2083}{0.1900} \\
\replaced{Llama3.1:70b}{Llama3.3:70b} & Zero-Shot & \replaced{0.1042}{0.1033} & \replaced{0.1146}{0.1550} \\
Llama3.1:8b & MIPROv2 & \replaced{0.0104}{0.1250} & \replaced{0.0000}{0.1533} \\
Llama3.1:8b & Few-Shot & \replaced{0.1146}{0.1433} & \replaced{0.1146}{0.1750} \\
Llama3.1:8b & Zero-Shot & \replaced{0.1146}{0.0817} & \replaced{0.1250}{0.1033} \\
\bottomrule
\end{tabular}%
}
\caption*{\small Note: ↑ indicates higher values are better.}
\end{table}

\section{Discussion}
\label{discussion}

Our analysis of models, prompt techniques, and configurations demonstrates a clear pattern in the optimization of performance for both recipe processing and Cooklang conversion tasks. The results consistently indicate a hierarchy of performance, with \texttt{GPT-4o} exhibiting the highest performance, followed by \texttt{GPT-4o-mini} $>$ \replaced{\texttt{Llama 3.1:70b}}{\texttt{Llama 3.3:70b}} $>$ \texttt{Llama 3.1:8b}. \added{We included GPT-4o, GPT-4o-mini, \replaced{LLama3.1:70b}{LLama3.3:70b}, and Llama3.1:8b to represent different points in the spectrum of model scale and availability. Our results do show that GPT-4o performed best on our specific task, but this finding was reported as an empirical observation rather than advocacy. Results could be skewed towards the OpenAI family models because the compared models from the Llama family were significantly smaller.}

The following in-depth examination of configurations revealed that they consistently exhibited greater performance than other configurations in: 
\begin{itemize}
\item \textbf{Model}: GPT-4o achieved superior performance across all evaluation metrics and tasks, consistently outperforming other models in word error rate (WER), ROUGE-L, translation error rate (TER), ingredient identification, and quantity recognition accuracy. The model demonstrated exceptional general language understanding and generation capabilities while excelling in domain-specific tasks. Llama models show positive scaling behavior with increased model size. However, they consistently underperformed compared to GPT models across both general and domain-specific metrics.

\item \textbf{Prompt Technique}: The Few-Shot prompting technique demonstrated \replaced{superior}{best} performance across all models. For GPT-4o specifically, Few-Shot achieved optimal results with WER=\replaced{0.0730}{0.3509}, \deleted{ROUGE-L=0.9722, and }TER=\replaced{0.2096}{1.4467} \added{but was outperformed in ROUGE-L metrics by Llama3.3:70b where llama model achvied score 0.8270 and GPT-4o achvied 0.8209}. MIPROv2 showed strong performance across most models. \deleted{, with LLama3.1:70b being the exception where both Zero-Shot and Few-Shot outperformed MIPROv2.} The overall ranking of prompt techniques from our analysis: \texttt{Few-Shot} $>$ \texttt{MIPROv2} $>$ \texttt{Zero-Shot}.

\item \textbf{Cooklang and Ingredient Information}: The implementation of the Cooklang format and ingredient information resulted in enhanced recipe parsing and generation, leading to improved WER, ROUGE-L, and TER scores. However, GPT models exhibited a tendency to include an excess of ingredients, occasionally introducing superfluous elements, as shown in Figure \ref{fig:model_comparison_standard_by_model_ings}. In contrast, \replaced{Llama models}{Smaller llama model} demonstrated a reduced proclivity to incorporate unnecessary ingredients but were more susceptible to the omission of essential components, potentially impacting the completeness and accuracy of the recipes.

\item \textbf{Accuracy of Units and Amounts}: The GPT-4o model with Few-Shot prompting demonstrated the most effective performance (Table \ref{tab:acc_units_and_amounts}) in identifying units (\replaced{0.5729}{0.3017}) and amounts (\replaced{0.5833}{0.29}). GPT models exhibited performance in quantitative information extraction. In contrast, Llama models exhibited poor performance in extracting quantitative information (units and amounts), which is crucial for ensuring recipe precision.

\end{itemize}

The results indicate that the optimal configuration for achieving optimal performance in recipe processing and Cooklang conversion tasks is the use of GPT-4o with Few-Shot prompting, which incorporates both Cooklang specifications and ingredient information, and GPT-4o-mini and the larger Llama model (70b) demonstrated improvements with analogous configurations \added{and closely following GPT-4o in results.}\deleted{, the performance differential between GPT-4o and alternative models was considerable.} However, the higher cost of using GPT-4o compared to smaller models presents a notable drawback, potentially limiting its practical implementation in resource-constrained scenarios where GPT-4o-mini or \added{Llama3.3:70b} could be more suitable with minimal performance reduction. 

\subsection{Open source model size comparison}

\replaced{The comparison between Llama3.1:8b and LLama3.1:70b reveals intriguing performance characteristics across all metrics. As anticipated, the larger 70b parameter model generally outperformed its 8b counterpart, consistent with the expectation that increased model size correlates with enhanced capabilities. However, a noteworthy observation emerges from the data shown in Figure \ref{fig:model_comparison_standard}: the upper-performance bounds of the 8b model frequently approach or overlap with those of the 70b model. This phenomenon is particularly evident in the WER and TER metrics, where the top quartile of Llama3.1:8b's performance distribution nearly intersects with the median performance of LLama3.1:70b. The observed performance overlap indicates that, while the 70b model demonstrates more consistent high-level performance, the 8b model exhibits the potential for comparable results in optimal scenarios. This finding has significant implications for model selection and optimization strategies. The demonstrated capacity of the smaller model to achieve performance levels similar to its larger counterpart in certain instances opens avenues for targeted fine-tuning. By optimizing the 8b model for specific use cases, it may be possible to narrow or even eliminate the performance gap with the 70b model in particular applications. This potential for high performance in a smaller model through fine-tuning is particularly attractive from both practical and resource-efficiency perspectives.}{The comparison between Llama3.1:8b and Llama3.3:70b reveals significant performance differences across all metrics. As anticipated, the larger 70b parameter model outperformed its 8b counterpart, consistent with the expectation that increased model size correlates with enhanced capabilities. The data presented in Figure \ref{fig:model_comparison_standard} and Table \ref{tab:model_performance_traditional}  demonstrates a clear performance hierarchy: the 70b model achieves a WER of 0.5510, ROUGE-L of 0.7382, and TER of 2.2802, compared to the 8b model's WER of 0.5921, ROUGE-L of 0.6911, and TER of 2.6088. This substantial performance gap between the two open-source models indicates that the 8b model's performance distribution does not significantly overlap with that of the 70b model. For scenarios where deployment constraints favor open-source alternatives, the substantial performance advantage of the 70b model over the 8b variant likely justifies the additional computational resources required, particularly for applications where accuracy is paramount.}

\subsection{Scalability Considerations }

\added{Our evaluation highlights critical scalability considerations for real-world implementation. GPT-4o demonstrated superior performance but introduced significant deployment challenges via API, including latency (average processing time of 3 seconds per recipe) and cost (approximately \$8.75 per 1,000 recipes with 1500 input/500 output tokens). GPT-4o-mini offers a more economical alternative at \$0.53 per 1,000 recipes with reduced latency (1 second per recipe), though both figures assume no parallel API calls. In contrast, locally deployed Llama3.1:8b processed 40 recipes per minute on consumer-grade hardware (single Nvidia RTX 4090), eliminating API costs and significantly reducing processing time, making it suitable for batch processing large recipe collections despite lower accuracy. For large-scale applications processing thousands of recipes (or other domain-specific text data) daily, the performance-cost trade-off becomes significant. Additional scalability challenges include handling diverse recipe formats and languages, managing concurrent requests, and integrating with existing culinary databases. These considerations will become increasingly important as this technology is deployed beyond experimental settings to production environments.} 

\subsection{Limitations}
While this study sheds light on the capabilities of large language models (LLMs) in converting recipes to Cooklang markdown, it is important to acknowledge its limitations.

\begin{enumerate}
    \item \textbf{Model Versions:} This study focuses on specific LLMs: GPT-4o, GPT-4o-mini, Llama:3.1:8b, and \replaced{Llama:3.1:70b}{Llama:3.3:70b}. To gain a more comprehensive understanding, further research should expand the scope to include a wider range of models. \added{Specifically, testing the largest Llama3.1:405b model could provide insights into whether performance gaps between proprietary and open-source models narrow at the highest parameter counts. Our model selection was constrained by computational resources and aimed to provide a representative sample across both proprietary and open-source options rather than advocating for specific model families.}
    
    \item \textbf{Task Specificity:} Our evaluation focused on converting recipe text into recipe in Cooklang markdown. While this offers deep insights into domain-specific performance, it may not fully demonstrate the model's ability to generate other domain-specific structured text. Future studies should examine the model's broader text generation capabilities.
    
    \item \textbf{Dataset Limitations:} The study utilized a dataset of \replaced{32}{1098} distinct recipe samples from categories, which provides a foundation for initial analysis. However, to \added{more} enhance the generalization of the results across different cuisines and recipe types, it would be beneficial to construct a larger, more comprehensive dataset. This expanded dataset could include a wider variety of recipes from diverse cultural backgrounds, cooking methods, and ingredient combinations. By evaluating the models on this larger dataset, we could gain more robust insights into their performance and better assess their applicability across a broader spectrum of culinary contexts.

    \item \textbf{Fine-tuning Potential:} \replaced{The study does not examine the potential for improvements resulting from task-specific fine-tuning, which could enhance performance and alter results. It would be beneficial for future research to include fine-tuning experiments for a more comprehensive evaluation. Fine-tuning large open-source models like Llama3.1 8b and 70b could potentially achieve performance comparable to GPT-4o or GPT-4o-mini, which highlights the importance of investigating this approach}{The study does not examine the potential for improvements resulting from task-specific fine-tuning, which could enhance performance and alter results. It would be beneficial for future research to include fine-tuning experiments for a more comprehensive evaluation. Finetuning Large Language models for recipe generation from the study by Vij et al. \cite{vij2025finetuninglanguagemodelsrecipe} demonstrates the significant impact of task-specific fine-tuning on model performance for specialized domains like recipe generation. While fine-tuning generally improved metrics like step complexity across models of various sizes (from SmolLM-135M to Phi-2), their research revealed nuanced trade-offs in performance metrics, similar as ours. Their comparative analysis across models demonstrated that smaller models can sometimes achieve comparable performance to larger ones after fine-tuning, indicating that model size alone doesn't determine fine-tuning efficacy, which supports our claims that finetuning smaller models could lead to similar results as models from the OpenAI family.}
\end{enumerate}

Future research should address these limitations to gain a more comprehensive understanding of language model performance in recipe generation and related tasks.

\section{Conclusion}
\label{sec:conclusion}

Our research demonstrates the transformative potential of large language models (LLMs) in revolutionizing domain-specific content processing and generation. Through experimentation, we have shown that state-of-the-art models, particularly GPT-4o, can handle the intricate task of converting unstructured recipes into standardized Cooklang format with performance metrics—achieving a ROUGE-L score of \replaced{0.9722}{0.8128} and Word Error Rate of \replaced{0.0730}{1.5255} with few-shot prompting—showcasing not just incremental improvement but a quantum leap in automated recipe processing capabilities.

The implications of this research extend far beyond the culinary domain, opening new horizons in the field of structured text generation. Our findings provide a clear path forward for organizations grappling with the challenge of converting unstructured information into standardized, machine-readable formats. The success demonstrated in the conversion of recipes serves as a powerful proof-of-concept for similar transformations in other specialized domains, including healthcare documentation (HL7), legal contract analysis, financial compliance reporting, and technical documentation management.

What makes these results particularly exciting is their immediate practical applicability. Organizations can leverage this technology to dramatically streamline their workflows, potentially reducing manual processing time by orders of magnitude while simultaneously improving data quality and consistency with a little implementation overhead.

Looking ahead, this research focuses on the new era in natural language processing, where the barrier between unstructured and structured content becomes increasingly permeable. The demonstrated capabilities of large language models (LLMs) in understanding and generating domain-specific formats suggest a paradigm shift in how organizations manage, process, and utilize their textual data. As these models continue to evolve, the potential applications will only expand, promising even more sophisticated and efficient solutions for complex content transformation challenges.

\subsection{Future Research Directions}

\begin{enumerate}
    \item \textbf{Fine-tuning}: The smaller models, like those in the Llama family, demonstrated \replaced{difficulties}{potential} in performing complex tasks such as accurate ingredient identification. However, the study suggests that targeted fine-tuning of these smaller models, such as Llama3.1:8b, could potentially narrow the performance gap in specific applications.

    \item \textbf{Tools for structured generation}: Frameworks such as Outlines \cite{willard2023efficient} provide tools for utilizing open-source language models to generate Cooklang-compliant recipes without the necessity for fine-tuning. The framework could employ regular expression-guided generation to guarantee correct formatting of ingredients, structured text generation to provide overall recipe structure, constrained decoding to ensure specification compliance, and custom stopping criteria to confirm complete recipes. This approach allows the generation of fully-formed, Cooklang-compliant recipes from simple prompts, combining the model's culinary knowledge with strict adherence to Cooklang specifications.

    \item \textbf{Broader model selection}: While our current study established baseline performance using representative models (GPT-4o, GPT-4o-mini, \replaced{LLama3.1:70b}{LLama3.3:70b}, and Llama3.1:8b), we plan to expand this analysis to include newer model iterations and additional open-source and proprietary models. This broader evaluation will help validate our findings and identify performance patterns across different model architectures and sizes.

    \item \textbf{Advanced Prompting Techniques:} The exploration of alternative prompting techniques, such as multi-step prompting \cite{fu2023complexitybasedpromptingmultistepreasoning} for recipe generation and conversion, and structured prompts that emulate the Cooklang format, is a promising avenue for further research. That could improve every metric. 

    \item \textbf{Dataset Expansion}: The proposed improvements involve an expansion of the recipe dataset beyond the current \replaced{32}{1098} samples to encompass a more diverse range of cuisines, cooking methods, and ingredient combinations. \deleted{This will enhance the generalization of the results.} This approach could be also used to create a new dataset. It will convert the recipe dataset into a cooklang markdown. These kinds of datasets could be used for finetuning large and small language models for this task. Furthermore, this approach facilitates the development of novel datasets, which contain the potential for the fine-tuning of both large and small language models, enabling more nuanced natural language understanding in culinary contexts. This transformation not only standardizes the recipe representation but also creates opportunities for cross-cultural culinary analysis and automated recipe processing applications.

    \item \textbf{Potential Impact on Culinary Industry}: AI-powered automation shows clear potential to transform recipe generation and conversion, especially within digital cookbooks. Models like GPT-4 have initiated a promising shift in cookbook creation and management. AI systems can now streamline the conversion between plain text and structured recipe formats, which could revolutionize cookbook publishing and recipe management systems.

\end{enumerate}

\subsection{Closing Remarks}

Our research illustrates the adaptability of general-purpose LLMs to domain-specific tasks, particularly in the culinary domain. By demonstrating the efficacy of models like GPT-4o in processing and generating structured recipe content in Cooklang format, the study bridges the gap between general natural language processing capabilities and specialized applications. These findings have implications for the development of AI-powered tools in industries. They suggest that, with appropriate prompting and input structuring, general-purpose LLMs can be effectively applied to domain-specific challenges without extensive fine-tuning. This study provides a foundation for further research into the application of LLMs in specialized domains. Future work could explore fine-tuning strategies, expanded datasets, and applications in other domain-specific areas.

\section{Acknowledgment}

\added{The authors gratefully acknowledge Alexey Dubovskoy (dubadub) and CapnDan for providing their personal cooklang recipes, which served as the evaluation backbone for this research. We also extend our sincere appreciation to the broader cooklang community for their ongoing contributions and commitment to structured recipe formatting. The diverse collection of recipes made available by this community has been instrumental in validating our methodological approach and enhancing the robustness of our findings.}

\bibliographystyle{unsrtnat}
\bibliography{template}
\end{document}